\documentclass[runningheads,a4paper]{llncs}

\usepackage{amssymb}
\usepackage{graphicx}
\usepackage{subfig}

\usepackage{url}
\urldef{\mailsa}\path|{alfred.hofmann, ursula.barth, ingrid.haas, frank.holzwarth,|
\urldef{\mailsb}\path|anna.kramer, leonie.kunz, christine.reiss, nicole.sator,|
\urldef{\mailsc}\path|erika.siebert-cole, peter.strasser, lncs}@springer.com|    
\newcommand{\keywords}[1]{\par\addvspace\baselineskip
\noindent\keywordname\enspace\ignorespaces#1}

\begin{document}

\mainmatter  

\title{Colon Shape Estimation Method for Colonoscope Tracking using \\Recurrent Neural Networks}

\titlerunning{Colon Shape Estimation Method for Colonoscope Tracking using RNN}

\author{Masahiro Oda\inst{1}%
\and 
Holger R. Roth\inst{1}
\and 
Takayuki Kitasaka\inst{2}
\and 
Kasuhiro Furukawa\inst{3}
\and 
Ryoji Miyahara\inst{4}
\and 
Yoshiki Hirooka\inst{3}
\and 
Hidemi Goto\inst{4}
\and \\
Nassir Navab\inst{5}
\and 
Kensaku Mori\inst{1,6}
}
\authorrunning{Masahiro Oda et al.}

\institute{Graduate School of Informatics, Nagoya University, Nagoya, Japan\\
\email{moda@mori.m.is.nagoya-u.ac.jp}\\
\and
School of Information Science, Aichi Institute of Technology, Toyota, Japan\\
\and
Department of Endoscopy, Nagoya University Hospital, Nagoya, Japan\\
\and
Department of Gastroenterology and Hepatology, \\Nagoya University Graduate School of Medicine, Nagoya, Japan\\
\and 
Technical University of Munich, M\"{u}nchen, Germany\\
\and
Research Center for Medical Bigdata, \\National Institute of Informatics, Tokyo, Japan
}

\toctitle{Lecture Notes in Computer Science}
\tocauthor{**** ****}
\maketitle

\begin{abstract}
We propose an estimation method using a recurrent neural network (RNN) of the colon's shape where deformation was occurred by a colonoscope insertion.
Colonoscope tracking or a navigation system that navigates physician to polyp positions is needed to reduce such complications as colon perforation.
Previous tracking methods caused large tracking errors at the transverse and sigmoid colons because these areas largely deform during colonoscope insertion.
Colon deformation should be taken into account in tracking processes.
We propose a colon deformation estimation method using RNN and obtain the colonoscope shape from electromagnetic sensors during its insertion into the colon.
This method obtains positional, directional, and an insertion length from the colonoscope shape.
From its shape, we also calculate the relative features that represent the positional and directional relationships between two points on a colonoscope.
Long short-term memory is used to estimate the current colon shape from the past transition of the features of the colonoscope shape.
We performed colon shape estimation in a phantom study and correctly estimated the colon shapes during colonoscope insertion with 12.39 (mm) estimation error.

\end{abstract}

\section{Introduction}


CT colonography (CTC) is currently performed as one of methods to find colonic polyps from CT images.
If colonic polyps or early-stage cancers are found in a CTC, a colonoscopic examination or polypectomy is performed to endoscopically remove them.
During a colonoscopic examination, a physician controls the colonoscope based on its camera view.
However, its viewing field is unclear because the camera is often covered by fluid or the colonic wall.
Furthermore, the colon changes shape significantly during colonoscope insertion.
Physicians require great experience and skill to estimate how the colonoscope is traveling inside the colon.
Inexperienced physicians overlook polyps or such complications as colon perforation.
A colonoscope navigation system is needed that leads a physician to the polyp position.
To develop a colonoscope navigation system, a colonoscope tracking method must be developed.

Endoscope tracking methods have been proposed by several research groups \cite{peters08,deligianni05,rai08,deguchi09,gildea06,schwarz06,liu13,ching10,fukuzawa15,oda17}.
For bronchoscope tracking, image- and sensor-based methods exist.
Image-based methods estimate the camera positions and movements based on 2D/3D image registrations.
Registrations between temporally continuous bronchoscopic images \cite{peters08} or between real and virtualized bronchoscopic images \cite{deligianni05,rai08,deguchi09} are used for tracking.
Sensor-based tracking methods use small position and direction sensors attached to a bronchoscope \cite{gildea06,schwarz06}.
For colonoscope tracking, image- and sensor-based methods also exist.
The image-based method \cite{liu13} has difficulty continuing to track when unclear colonoscopic views are obtained.
Electromagnetic (EM) sensors are used to obtain colonoscope shapes \cite{ching10,fukuzawa15}.
Unfortunately, they cannot guide physicians to polyp positions because they cannot map the colonoscope shape to a colon in a CT volume, which may contain polyp detection results.
A colonoscope tracking method that uses CT volume and EM sensors was reported \cite{oda17}.
It obtains two curved lines that representing the colon and colonoscope shapes to estimate the colonoscope position on a CT volume coordinate system.
This method enables real-time tracking regardless of the colonoscopic image quality.
However, this method does not consider the colon deformations caused by colonoscope insertions.
Large tracking errors were observed at the transverse and sigmoid colons, which are significantly deformed by a colonoscope insertion.
To improve the tracking accuracy, we need to develop a method that estimates the colon shape during colonoscope insertions.

We propose a method that estimates the colon shape with the deformations caused by colonoscope insertion.
The shape of the colonoscope, which is inserted into the colon, affects the colon's deformation.
We propose a shape estimation network (SEN) to model the relationships between the colon and colonoscope shapes by a deep learning framework.
After training, SEN estimates the colon shape from the colonoscope shape.
SEN has a long short-term memory (LSTM) layer \cite{hochreither97}, which is a recurrent neural network (RNN), to perform estimations based on temporal transitions.
To make maximum use of the colonoscope shape information, we developed a relative feature of the shape.
Relative, positional, and directional features are given to SEN for the estimations.
We performed a phantom study to confirm the performance of the proposed method.

The followings are the contributions of this paper: {\bf (1)} it propose a new deep learning framework that models the relationships between the organ shape and the forces that cause organ deformations and {\bf (2)} it introduce a new relative feature that represents 3D shape information as a 2D matrix shape. The feature can be processed by convolutional neural networks (CNNs) to extract features.

\section{Colon Shape Estimation Method}

\subsection{Overview}

We estimate the colon shape from the colonoscope shape.
These shapes are the temporal information that was observed during the colonoscope insertions.
The estimation is performed using a SEN with CNN and LSTM layers.
We extract the relative features of the colonoscope shape using CNN layers and combine them with other features that are processed by a LSTM layer.
LSTM performs regression based on the temporal transition of the feature values.

\subsection{Colon and colonoscope shape representation}

\begin{figure}[tb]
  \begin{minipage}{0.15\textwidth}
    \centering
\subfloat[]{\includegraphics[width=0.7\textwidth, clip, trim=0 340 850 0]{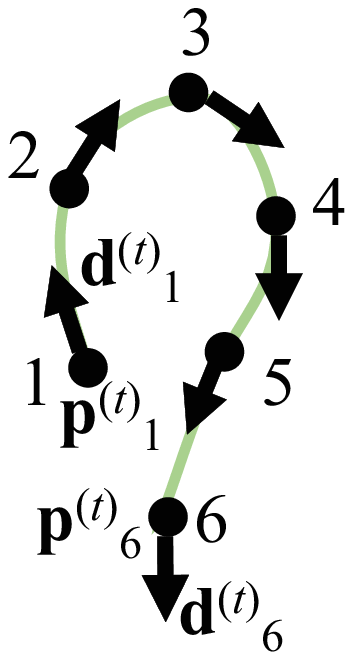}\label{fig:featurecalculation1}}
  \end{minipage}
  \begin{minipage}{0.15\textwidth}
    \centering
\subfloat[]{\includegraphics[width=0.7\textwidth]{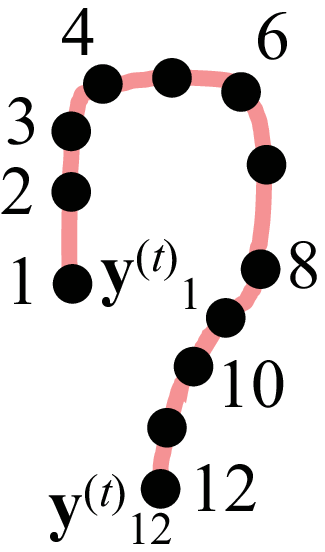}\label{fig:featurecalculation2}}
  \end{minipage}
  \begin{minipage}{0.7\textwidth}
\centering
\subfloat[]{\includegraphics[width=0.75\textwidth, clip, trim=180 300 100 10]{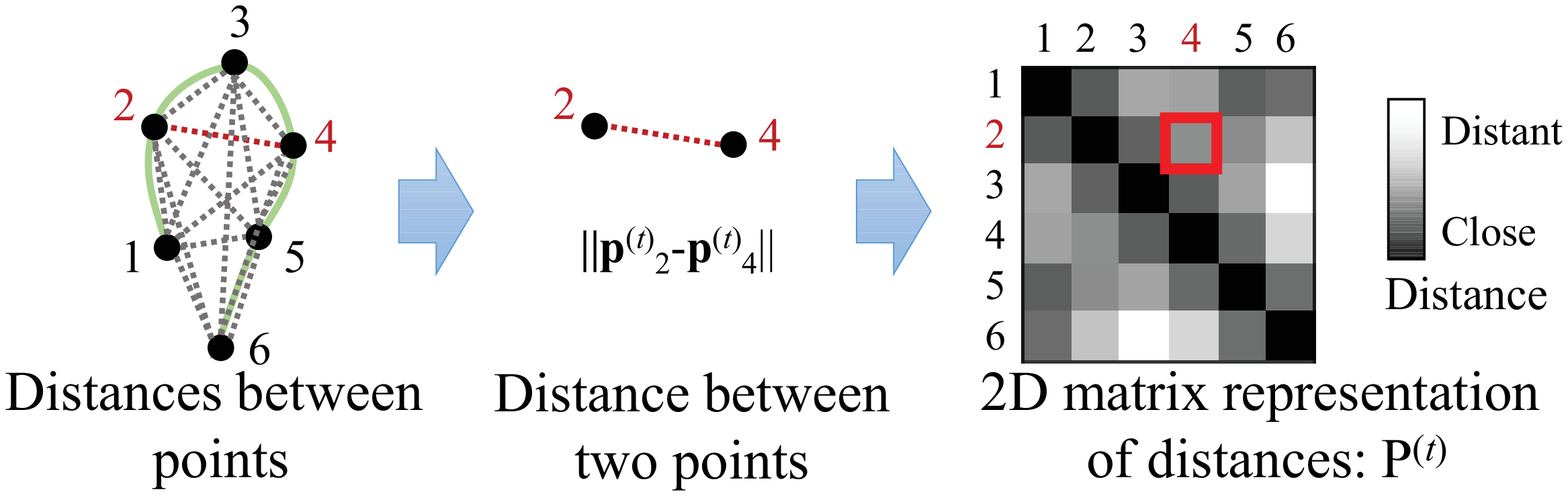}\label{fig:featurecalculation3}}\\
\subfloat[]{\includegraphics[width=0.75\textwidth]{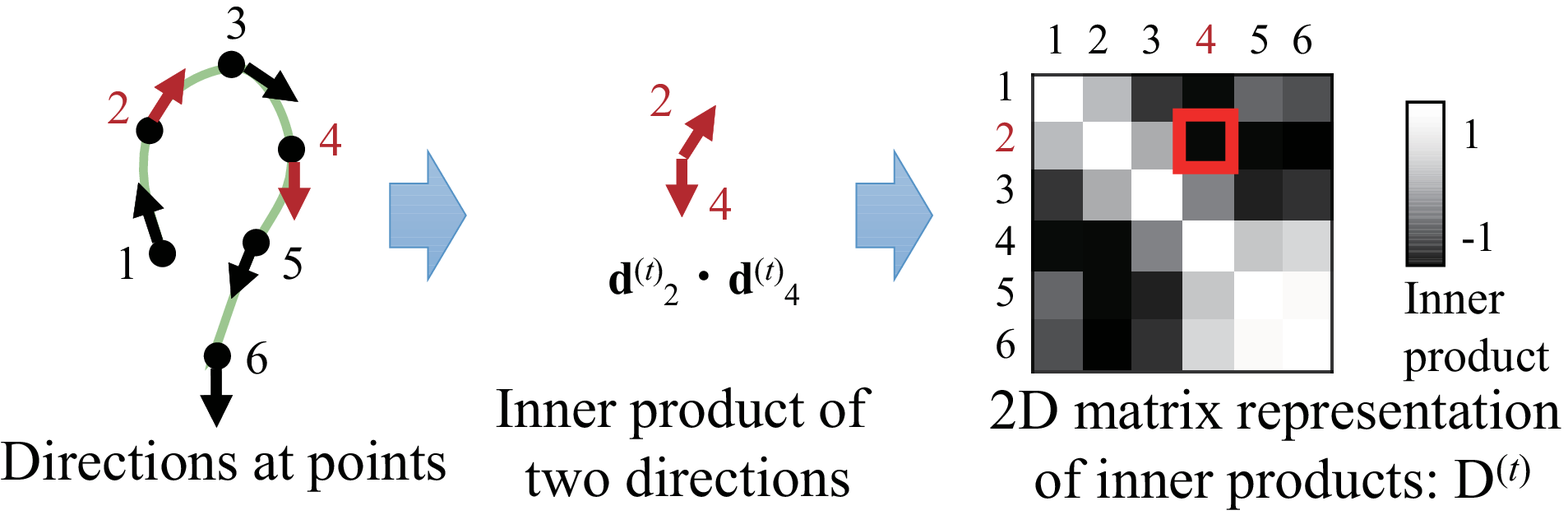}\label{fig:featurecalculation4}}
  \end{minipage}\hfill
\caption[]{(a) Green curved line represents colonoscope fiber shape. 
(b) shows ${\bf y}_{m}^{(t)}$ on colon centerline. (c) shows positional relation feature. Distances between two points are stored in 2D-matrix ${\rm P}^{(t)}$. (d) shows directional relation feature. Inner products of two directions are stored in 2D-matrix ${\rm D}^{(t)}$.}\label{fig:featurecalculation}
\end{figure}

We used a point-set representation to describe the colonoscope and colon shapes.
Both are represented as sets of points aligned along the colonoscope and colon centerlines.
The colonoscope shape of time $t \ (t=1, \ldots, T)$ is a set of points and directions ${\bf X}^{(t)} = \{ {\bf p}_{n}^{(t)}, {\bf d}_{n}^{(t)}; \ n=1, \ldots, N \}$ related to the colonoscope (Fig. \ref{fig:featurecalculation} (a)).
${\bf p}_{n}^{(t)}$ is a point aligned along the colonoscope.
${\bf d}_{n}^{(t)}$ is a tangent direction of the colonoscope fiber at ${\bf p}_{n}^{(t)}$.
$T$ is the total number of time frames and $N$ is the total number of the points in the colonoscope shape.
The colon shape is a set of points ${\bf Y}^{(t)} = \{ {\bf y}_{m}^{(t)}; \ m=1, \ldots, M \}$ aligned along a colon centerline (Fig. \ref{fig:featurecalculation} (b)).
$M$ is the total number of points in the colon shape.

\subsection{Colonoscope shape features}

From ${\bf X}^{(t)}$, we calculate the features that related to the colonoscope shape.

\subsubsection{Structure features}
Structure feature ${\bf A}^{(t)}$ includes ${\bf p}_{n}^{(t)}$, ${\bf d}_{n}^{(t)}$, and the insertion length of colonoscope $l^{(t)}$, calculated as follows.
We applied the Hermite spline interpolation \cite{chen10} to generate a curved line that is connected to ${\bf p}_{n}^{(t)}$.
The curved line's length is used as the insertion length of colonoscope $l^{(t)}$.
A structure feature is a set of these values ${\bf A}^{(t)} = \{ {\bf p}^{(t)}_{n}, {\bf d}^{(t)}_{n}, l^{(t)} \}$.

\subsubsection{Relative features}
Relative features include the positional relations between pairs of ${\bf p}_{n}^{(t)}$ and the directional relations between pairs of ${\bf d}_{n}^{(t)}$.
Positional relation feature ${\rm P}^{(t)}$ is a $N \times N$ matrix with the distances between ${\bf p}_{i}^{(t)}$ and ${\bf p}_{j}^{(t)} \ (i, j=1, \ldots, N)$ as $(i, j)$ elements (Fig. \ref{fig:featurecalculation} (c)).
Directional relation feature ${\rm D}^{(t)}$ is a $N \times N$ matrix with inner products ${\bf d}_{i}^{(t)} \cdot {\bf d}_{j}^{(t)}$ as $(i, j)$ elements (Fig. \ref{fig:featurecalculation} (d)).
${\rm P}^{(t)}$ and ${\rm D}^{(t)}$ contain positional and directional relationship information.

\subsection{Shape estimation network}

\begin{figure}[tb]
\begin{center}
\includegraphics[width=0.8\textwidth, clip, trim=0 295 0 0]{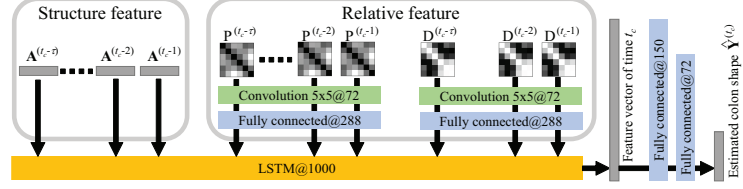}
\caption{Structure of shape estimation network (SEN). Input is colonoscope shape features in past time period $t=t_{c}-\tau, \ldots, t_{c}-1$. Output is estimated colon shape of current time $t_{c}$. Numbers written after @ are kernel or unit numbers.}
\label{fig:network}
\end{center}
\end{figure}

We designed SEN with colonoscope shape feature input paths and the output of colon shape parameters (Fig. \ref{fig:network}).
Among the shape features, the relative features are processed by convolutional layers, which analyze the positional and directional relationship of the points on the colonoscope shape.
The features extracted by the convolutional layers are combined with the structure features and given to a LSTM layer, which considers the temporal transition of all the features.
To perform estimation utilizing temporal information, the SEN input is the features in a past time period $t=t_{c}-\tau, \ldots, t_{c}-1$ until current time $t_{c}$.
The LSTM layer's output is processed by fully connected layers.
The final layer outputs estimated colon shape $\hat{{\bf Y}}^{(t_{c})}$.

\section{Experimental Setup}

We confirmed the colon shape estimation performance of our method in phantom-based experiments.
Our method needs pairs of ${\bf X}^{(t)}$ and ${\bf Y}^{(t)}$ at every time step for training SEN.
${\bf X}^{(t)}$ and ${\bf Y}^{(t)}$ were measured using an EM and distance sensors.
We used a colon phantom (colonoscopy training model type I-B, Koken, Tokyo, Japan), a CT volume of the phantom, a colonoscope (CF-Q260AI, Olympus, Tokyo, Japan), an EM sensor (Aurora 5/6DOF Shape Tool Type 1, NDI, Ontario, Canada), and a distance image sensor (Kinect v2, Microsoft, WA, USA).

In colonoscopic examinations, physicians observe and treat the colon while retracting the colonoscope after its insertion up to the cecum.
We assume the colonoscope tip is inserted up to the cecum when the colonoscope tracking starts.
The proposed colon shape estimation method is also used during colonoscope tracking.
The colonoscope was moved from the cecum to the anus.

\subsection{Colonoscope shape measurement}

The EM sensor is strap-shaped with six sensors at its tip and points along its strap-shaped body.
Each sensor gives the 3D position and the 3D/2D direction along the colonoscope by inserting the sensor into the colonoscope working channel.
The measured data are a set of points and directions ${\bf X}^{(t)} = \{ {\bf p}_{n}^{(t)}, {\bf d}_{n}^{(t)}; \ n=1, \dots, 6 \}$ at time $t$.
They are used as the colonoscope shape.

\subsection{Colon shape measurement}

We used a 3D printer to make 12 position markers to detect the surface position of the colon phantom, which has an easy-to-detect color and shape.
The blue marker gives good color contrast to the orange colon phantom.
The marker has a spherical shape, which enables detection from all directions.
The position markers are attached to the surface of the colon phantom.

The distance image sensor is mounted to measure the surface shape of the colon phantom (Fig. \ref{fig:measurement}).
We obtained both distance and color images from the sensor.
We applied an automated marker position extraction process to these images to obtain 12 three-dimensional points of the markers.
The measured points of the markers were aligned along the colon centerline and numbered.
The colon centerline was extracted from the CT volume of the colon phantom.
The numbered markers are described as ${\bf Y}^{(t)} = \{ {\bf y}_{m}^{(t)}; \ m=1, \ldots, 12 \}$ at time $t$.
${\bf y}_{1}^{(t)}$ and ${\bf y}_{12}^{(t)}$ respectively correspond to markers near the cecum and the anus.
${\bf Y}^{(t)}$ is the colon shape.

\subsection{Shape estimation network training}

\begin{figure}[tb]
\begin{center}
\includegraphics[width=0.75\textwidth]{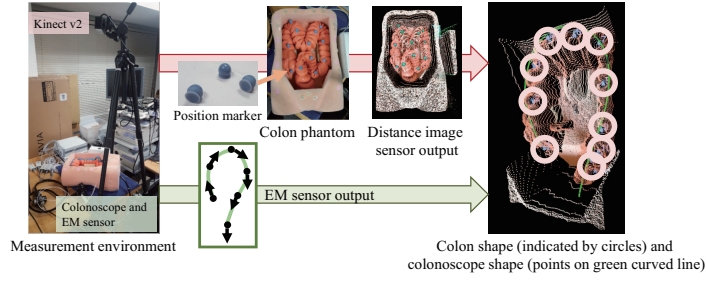}
\caption{Colonoscope and colon shapes measurement setup: Colon phantom and sensors are mounted, as shown on the left. 
}
\label{fig:measurement}
\end{center}
\end{figure}

We simultaneously recorded both ${\bf X}^{(t)}$ and ${\bf Y}^{(t)}$ during colonoscope insertions to the phantom.
The measurements were performed using the experimental setup shown in Fig. \ref{fig:measurement}.
The shapes were recorded six times per second.
Inaccurate measurement results caused by the mis-detection were manually corrected.

${\bf X}^{(t)}$ and ${\bf Y}^{(t)}$ belong to the EM and distance image sensor coordinate systems.
We registered them in the CT coordinate system using the iterative closest point (ICP) algorithm \cite{besl92} and manual registrations.
Registered ${\bf X}^{(t)}$ and ${\bf Y}^{(t)}$ in the CT coordinate system were used to train the SEN under these conditions: $\tau = 20$ past frames used by the LSTM layer, 50\% dropout of fully connected layers, 50 mini batch size, and 480 training epochs.

\subsection{Colon shape estimation}

We measured the colonoscope shapes in a past time period, ${\bf X}^{(t_{c}-\tau)}, \ldots, {\bf X}^{(t_{c}-1)}$, during a colonoscope insertion.
These colonoscope shapes were registered to the CT coordinate system using the ICP algorithm and input to the SEN to obtain estimated colon shape $\hat{{\bf Y}}^{(t_{c})}$ of current time $t_{c}$.

\subsection{Evaluation metric}

We use the mean distance (MD) (mm) between ${\bf Y}^{(t)}$ and $\hat{{\bf Y}}^{(t)}$ as an evaluation metric.
The MD of one colonoscope insertion is described as
\begin{equation}
E = \frac{1}{12(T-\tau)} \sum^{T}_{t=\tau+1} \sum^{12}_{m=1} |\hat{{\bf y}}^{(t)}_{m} - {\bf y}^{(t)}_{m}|.
\end{equation}
This metric indicates how an estimated colon shape is close to a ground truth.

\section{Experimental Results}

\begin{figure}[tb]
\begin{center}
\begin{tabular}{ccc}
\includegraphics[width=0.5\textwidth]{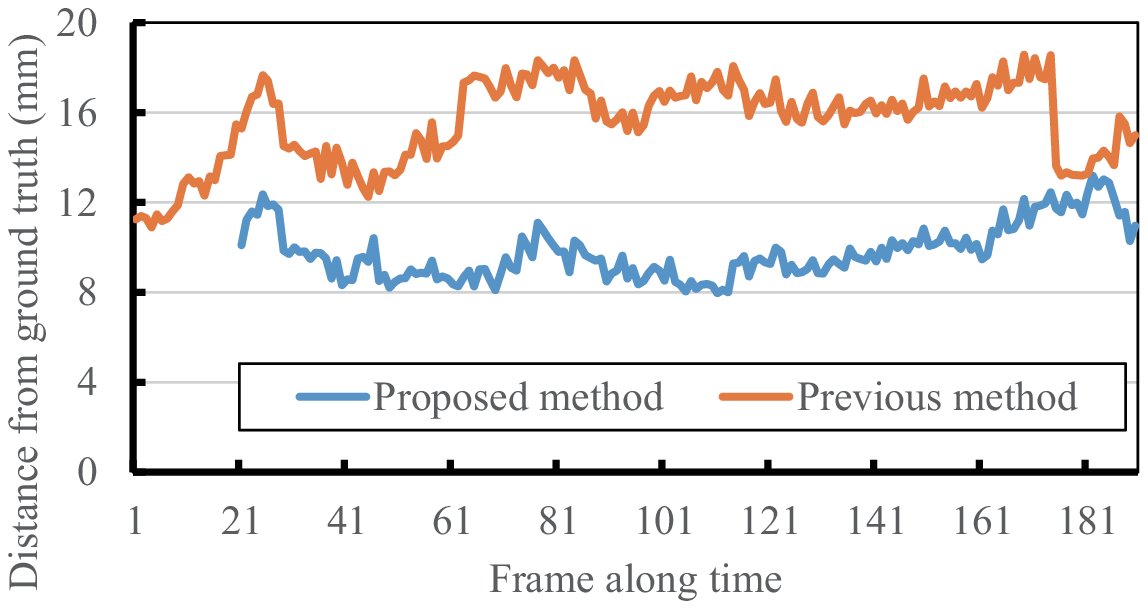} & 
\includegraphics[width=0.19\textwidth]{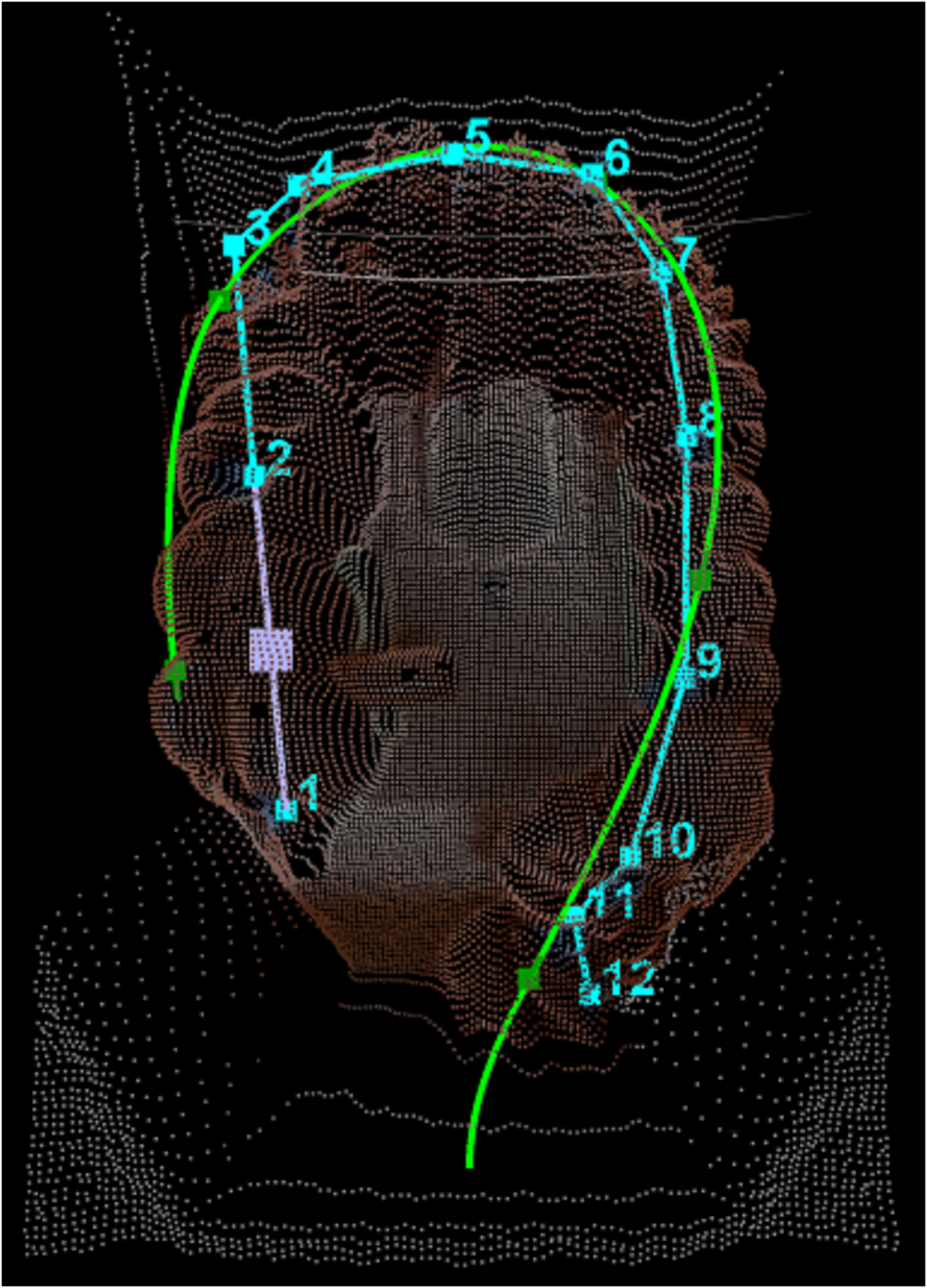} & 
\includegraphics[width=0.19\textwidth]{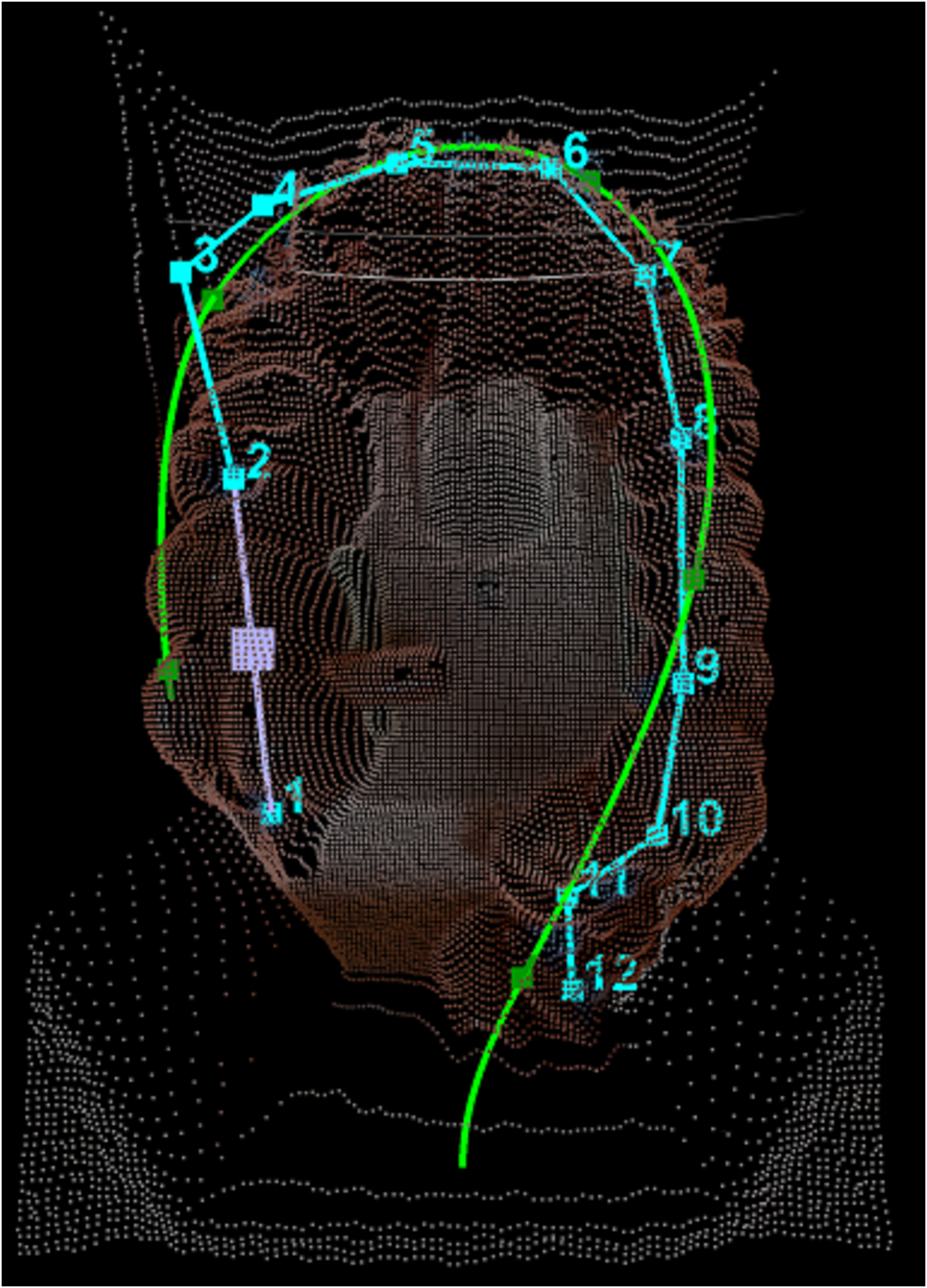} 
\\
(a) & (b) & (c) 
\end{tabular}
\end{center}
\caption{(a) is distances between ground truth and estimated colon shapes in each frame for proposed and previous \cite{oda18} methods. Proposed method starts estimation after $\tau = 20$ frames given. (b) and (c) show colonoscope shapes (points on green curved lines), estimated colon shapes (blue numbered points), and surface shapes of colon phantom (small dots). (b) and (c) are results of proposed and previous \cite{oda18} methods.}
\label{fig:result}
\end{figure}

We evaluated the following three colon shape estimation methods: (1) our proposed method, (2) the proposed method without a relative feature, and (3) the previous method \cite{oda18}.
The method \cite{oda18} estimates the colon shape from the colonoscope shape using regression forests and without temporal information.
We recorded colonoscope and colon shapes during seven colonoscope insertions and recorded 1,179 shapes.
An engineering researcher operated the colonoscope.
Shapes of six colonoscope insertions were used as training data, and the remaining colonoscope insertion was used as testing data.
We performed a leave-one-colonoscope-insertion-out cross validation in our evaluation.
The following are the MDs of the methods: (1) 12.39, (2) 12.61, and (3) 21.41 (mm).
The proposed method performed estimation with less error than the previous method (comparing (1) and (3)).
Also, using the relative feature reduced the errors (comparing (1) and (2)).
For methods (1) and (3), we compared distances between the ground truth (measured) and estimated colon shapes in each frame in Fig. \ref{fig:result} (a).
Estimation results of the proposed method were close to the ground truth in most of the frames.
Estimated colon shapes are shown in Figs. \ref{fig:result} (b) and (c).
The shape obtained from the proposed method was similar to the ground truth.

\section{Discussion}

The proposed SEN accurately and stably estimated the colon shape during colonoscope insertion.
SEN utilizes not only sensor information but also relative information and temporal transition for its estimation.
These features contributed to the improvement of the estimation accuracy.
Estimation results can be used to improve the colonoscope tracking accuracy \cite{oda17}.
The results of the proposed methods are important to achieve practical colonoscope tracking methods and will also contribute to the assistance of endoscopic procedures.

Our experimental result showed one application of the proposed method, which can be used as a soft organ shape estimation method of the forces that affect organ deformation.
For example, bronchus shape estimation during a bronchoscopic insertion and estimation of the organ deformation were caused by contact with surgical tools.
The proposed method models the relationships between the forces and organ deformations caused by the forces.
This modeling framework is applicable for many computer-assisted intervention topics.
The proposed method has a potential to work on phantom and real colons, even on colonoscope operations made by different operators.

The proposed SEN can be applied to estimate human colon shapes.
To do this, we need pairs of colon and colonoscope shapes during colonoscope insertions into human colons.
Taking X-ray images of the abdominal region is one candidate to observe these shapes, which we believe that we can extract from such X-ray images.
Once SEN is trained using human data, it estimates the colon shape.
This will enable colonoscope navigation during polypectomy.


This paper proposed a colon shape estimation method using an RNN technique.
SEN models the relationships between the colonoscope and colon shapes during colonoscope insertions.
SEN input includes the structure and relative features of colonoscope shapes.
SEN was trained to output a colon shape from these features.
We applied the proposed method to estimate colon phantom shapes.
The proposed method achieved more accurate and stable estimation results than the previous method.
Future work includes applications to a colonoscope tracking method and estimations of the human colon shape.

\subsubsection*{Acknowledgments.} 
Parts of this research were supported by the MEXT, the JSPS KAKENHI Grant Numbers 26108006, 17H00867, the JSPS Bilateral International Collaboration Grants, and the JST ACT-I (JPMJPR16U9).

\end{document}